\documentclass[10pt,onecolumn]{article}

\usepackage[margin=0.75in]{geometry}
\usepackage{float}
\usepackage{times}
\usepackage{amsmath}
\usepackage{amsfonts}
\usepackage{booktabs}
\usepackage{array}
\usepackage{longtable}
\usepackage{url}
\usepackage{graphicx}
\usepackage{caption}
\usepackage{balance}
\usepackage{natbib}

\usepackage{listings}
\usepackage{xcolor}
\definecolor{codegreen}{rgb}{0,0.6,0}
\definecolor{codegray}{rgb}{0.5,0.5,0.5}
\definecolor{codepurple}{rgb}{0.58,0,0.82}

\lstdefinelanguage{yaml}{
  keywords={true,false,null,yes,no},
  keywordstyle=\color{codepurple}\bfseries,
  sensitive=false,
  comment=[l]{\#},
  commentstyle=\color{codegreen}\ttfamily,
  stringstyle=\color{codepurple}\ttfamily,
  morestring=[b]',
  morestring=[b]"
}

\lstdefinelanguage{json}{
    basicstyle=\small\ttfamily,
    numbers=left,
    numberstyle=\tiny,
    stepnumber=1,
    numbersep=8pt,
    showstringspaces=false,
    breaklines=true,
    frame=lines,
    string=[s]{"}{"},
    stringstyle=\color{blue},
    comment=[l]{:},
    commentstyle=\color{black},
}

\usepackage{titling}
\pretitle{\begin{center}\LARGE\bfseries}
\posttitle{\par\end{center}\vskip 0.5em}
\preauthor{\begin{center}\large}
\postauthor{\end{center}}
\predate{\begin{center}\large}
\postdate{\par\end{center}\vskip 1em}

\usepackage{titlesec}
\titleformat{\section}{\large\bfseries}{\thesection}{1em}{}
\titleformat{\subsection}{\normalsize\bfseries}{\thesubsection}{1em}{}
\titleformat{\subsubsection}{\normalsize\bfseries}{\thesubsubsection}{1em}{}

\renewenvironment{abstract}
{\small\quotation\noindent\textbf{Abstract}\par}
{\endquotation}

\begin{document}

\title{How Much Do LLMs Hallucinate in Document Q\&A Scenarios?\\[0.3em]
{\large A 172-Billion-Token Study Across Temperatures, Context Lengths, and Hardware Platforms}}

% Custom column type for tables
\newcolumntype{L}[1]{>{\raggedright\arraybackslash}p{#1}}
\newcolumntype{C}[1]{>{\centering\arraybackslash}p{#1}}

\author{JV Roig\\
\small Kamiwaza AI\\
\small \texttt{jv@kamiwaza.ai}
}

\date{March 2026}

\maketitle

\begin{abstract}
\noindent
How much do large language models actually hallucinate when answering questions grounded in provided documents? Despite the critical importance of this question for enterprise AI deployments, reliable measurement has been hampered by benchmarks that rely on static datasets vulnerable to contamination, LLM-based judges with documented biases, or evaluation scales too small for statistical confidence. We address this gap using RIKER, a ground-truth-first evaluation methodology that enables deterministic scoring without human annotation. Across 35 open-weight models, three context lengths (32K, 128K, and 200K tokens), four temperature settings, and three hardware platforms (NVIDIA H200, AMD MI300X, and Intel Gaudi3), we conducted over 172 billion tokens of evaluation---an order of magnitude beyond prior work. Our findings reveal that: (1) even the best-performing models fabricate answers at a non-trivial rate---1.19\% at best at 32K, with top-tier models at 5--7\%---and fabrication rises steeply with context length, nearly tripling at 128K and exceeding 10\% for all models at 200K; (2) model selection dominates all other factors, with overall accuracy spanning a 72-percentage-point range and model family predicting fabrication resistance better than model size; (3) temperature effects are nuanced---T=0.0 yields the best overall accuracy in roughly 60\% of cases, but higher temperatures reduce fabrication for the majority of models and dramatically reduce coherence loss (infinite generation loops), which can reach 48$\times$ higher rates at T=0.0 versus T=1.0; (4) grounding ability and fabrication resistance are distinct capabilities---models that excel at finding facts may still fabricate facts that do not exist; and (5) results are consistent across hardware platforms, confirming that deployment decisions need not be hardware-dependent. We release all experimental data, including raw model outputs, to support reproducibility.
\end{abstract}

\section{Introduction}\label{sec:intro}

One of the most common and critical applications of large language models (LLMs) in the enterprise is answering questions grounded in provided documents. Whether through context stuffing, retrieval-augmented generation (RAG), or agentic retrieval, the core task is the same: given a set of documents, answer questions accurately based on what is in them --- and only what is in them.

The fundamental concern with all of these approaches is hallucination. When an LLM fabricates information that does not exist in the provided documents, or misattributes facts across documents, the consequences for enterprise deployments range from embarrassing to catastrophic. Yet despite the centrality of this concern, a simple question remains surprisingly difficult to answer: \textit{how much do LLMs actually hallucinate in document Q\&A scenarios?}

The difficulty lies not in the question but in the measurement. Existing approaches to quantifying hallucination suffer from three fundamental problems. First, static benchmarks are vulnerable to data contamination --- models may have seen the test data during training, inflating apparent performance. Second, LLM-as-judge approaches, while scalable, exhibit systematic biases that undermine reliability. Third, most evaluations operate at scales too small for statistical confidence, reporting single-run results on limited question sets.

In prior work, we introduced RIKER (Retrieval Intelligence and Knowledge Extraction Rating) \cite{roig2026scalablereliableevaluationai}, a methodology that addresses these challenges through paradigm inversion: generating documents \textit{from} known ground truth rather than extracting ground truth \textit{from} documents. This approach enables deterministic scoring at arbitrary scale, without human annotation, reference models, or LLM judges. Our initial RIKER study evaluated 33 models using over 21 billion tokens, establishing that context length claims frequently exceed usable capacity, that cross-document aggregation is substantially harder than single-document extraction, and that grounding ability and hallucination resistance are distinct capabilities.

In this work, we apply the RIKER methodology at significantly expanded scale to directly address the question in our title. The key extensions beyond the original RIKER study are:

\begin{itemize}
    \item \textbf{Massively expanded evaluation scale.} We increase from 21 billion to over 172 billion tokens of evaluation, with significantly more questions per context length to improve statistical reliability.
    \item \textbf{Systematic temperature analysis.} We evaluate every model at four temperature settings (0.0, 0.4, 0.7, and 1.0) with 8 runs each, enabling rigorous analysis of temperature effects on hallucination rates.
    \item \textbf{Cross-hardware validation.} We run identical experiments across NVIDIA H200, AMD MI300X, and Intel Gaudi3 platforms to determine whether hardware choice affects model behavior.
    \item \textbf{Updated model roster.} We include newer models not available during the original study, providing current rankings across multiple model families and sizes.
\end{itemize}

Our findings paint a sobering picture. Even under optimal conditions, every model fabricates answers at a non-trivial rate, and fabrication rises steeply with context length. We also find that the conventional wisdom of setting temperature to zero for factuality in enterprise document Q\&A scenarios is not universally supported---while T=0.0 yields the best overall accuracy in a majority of cases, it can increase both fabrication rates and coherence loss---alongside a fundamental decoupling between grounding ability and fabrication resistance, and confirmation that hardware platform does not meaningfully affect results.

The remainder of this paper is organized as follows. Section~\ref{sec:related} reviews related work on hallucination measurement and temperature effects. Section~\ref{sec:methodology} summarizes the RIKER methodology. Section~\ref{sec:design} describes our experimental design. Section~\ref{sec:results} presents results. Section~\ref{sec:discussion} discusses implications and limitations, and Section~\ref{sec:conclusion} concludes.

\section{Related Work}\label{sec:related}

A comprehensive review of knowledge retrieval evaluation, including long-context benchmarks, RAG evaluation frameworks, benchmark contamination, and synthetic data validity, is provided in the original RIKER paper \cite{roig2026scalablereliableevaluationai}. Here we focus on the three areas most directly relevant to this study: hallucination measurement, long-context evaluation, and the methodological challenges that motivate our approach.

\subsection{Hallucination Measurement}

Quantifying LLM hallucination remains an open challenge. TruthfulQA and SimpleQA \cite{wei2024measuringshortformfactualitylarge} evaluate factual accuracy, but face distinct challenges: TruthfulQA shows evidence of contamination \cite{deng2024investigatingdatacontaminationmodern}, while SimpleQA focuses on short-form responses where even frontier models struggle to achieve majority accuracy. FActScore \cite{min2023factscore} decomposes long-form outputs into atomic facts for verification. HaluEval \cite{li2023halueval} and HalluLens \cite{bang2025hallulensllmhallucinationbenchmark} provide additional hallucination benchmarks, while the Hugging Face Hallucinations Leaderboard \cite{huggingface2024hallucinations} aggregates results across multiple benchmarks. No benchmark adequately addresses long-form knowledge extraction from document corpora.

\subsection{Long-Context Evaluation}

The Needle-in-a-Haystack (NIAH) paradigm \cite{kamradt2023niah} places a random fact in a long context and tests retrieval, but fundamentally tests retrieval rather than comprehension. RULER \cite{hsieh2024rulerwhatsrealcontext} expands beyond retrieval to include multi-hop tracing, aggregation, and question answering; despite models claiming 32K+ context support, many fail to maintain satisfactory performance at that length. The ``Lost in the Middle'' phenomenon \cite{liu2024lostinthemiddle} demonstrates that LLMs struggle with information placed in the middle of long contexts. LongBench v2 \cite{bai2025longbenchv2deeperunderstanding} provides human-annotated tasks across six categories with contexts from 8K to 2M words, demonstrating that long-context comprehension remains challenging even for frontier models.

\subsection{Methodological Challenges}

Two pervasive problems affect LLM evaluation. First, data contamination: comprehensive surveys document how models trained on benchmark data produce inflated scores \cite{xu2024benchmarkdatacontaminationlarge, deng2024investigatingdatacontaminationmodern}, with simple paraphrasing sufficient to bypass decontamination measures \cite{yang2023rephrasedsamples}. Second, LLM-as-judge unreliability: the CALM framework identifies 12 distinct biases in LLM judges \cite{ye2024justiceprejudicequantifyingbiases}, and empirical validation of RAGAS metrics shows a harmonic mean correlation of only 0.55 with human evaluation \cite{beatrust2024ragasevaluation}.

RIKER \cite{roig2026scalablereliableevaluationai} addresses both problems through paradigm inversion --- generating documents from known ground truth --- enabling deterministic scoring at arbitrary scale without human annotation or LLM judges. The PICARD framework \cite{roig2025picard} provides the underlying contamination-resistant evaluation methodology. A more detailed treatment of the evaluation landscape, including retrieval benchmarks, multi-hop QA, RAG evaluation, GraphRAG, and synthetic data validity, is provided in the original RIKER paper. This study builds on RIKER to answer the specific question of how much LLMs hallucinate in document Q\&A, at a scale and across variables (temperature, hardware) not previously examined.

\section{Methodology}\label{sec:methodology}

This study uses the RIKER (Retrieval Intelligence and Knowledge Extraction Rating) methodology \cite{roig2026scalablereliableevaluationai}, which we summarize here. Full details, including the Coherent Simulated Universe approach, template-based document generation, and cross-corpus validation, are provided in the original RIKER paper.

\begin{figure}[H]
\centering
\includegraphics[width=\textwidth]{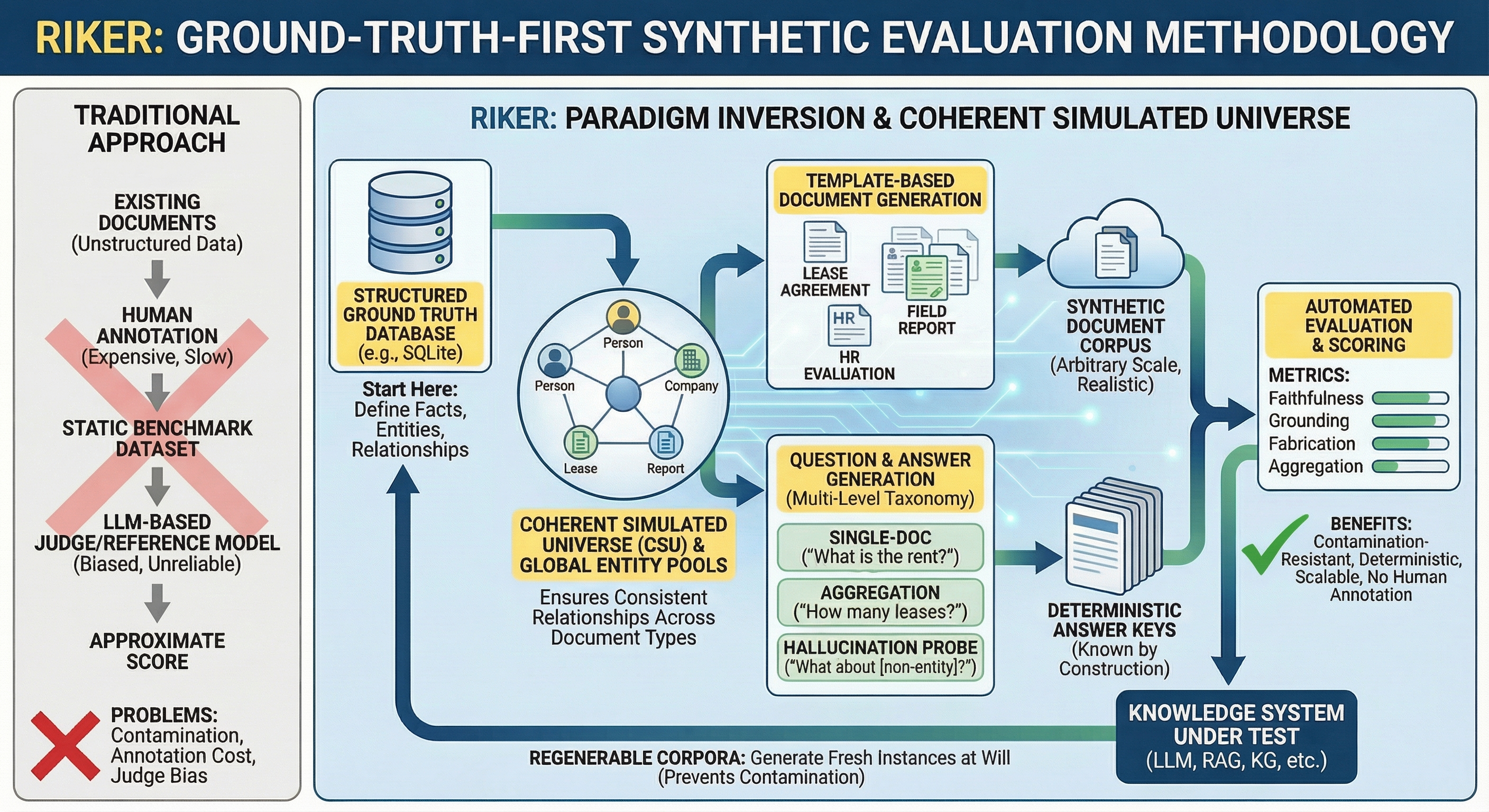}
\caption{RIKER methodology overview. Traditional approaches (left) extract ground truth from existing documents via expensive human annotation, producing static benchmarks vulnerable to contamination and requiring biased LLM judges. RIKER (right) inverts this: structured ground truth is defined first, then documents and questions are generated from it, enabling deterministic scoring and regenerable corpora. (Figure reproduced from \cite{roig2026scalablereliableevaluationai}.)}
\label{fig:riker-overview}
\end{figure}

RIKER inverts the traditional evaluation paradigm: rather than extracting ground truth from existing documents (requiring expensive human annotation), RIKER generates documents \textit{from} known ground truth. All entities, relationships, and facts are first populated in a relational database, and documents are then rendered as human-readable views of this underlying ground truth. This provides three key advantages: (1) every question has a verifiable answer by construction, enabling deterministic scoring; (2) corpora can be regenerated with different random seeds, enabling contamination resistance; and (3) the approach scales to arbitrary size without human annotation effort. Importantly, document generation is entirely template-based --- no LLM is involved in creating the corpus, avoiding the systematic biases that arise when LLMs generate their own benchmark data.

\subsection{Question Taxonomy}

RIKER generates questions across twelve difficulty levels organized into three categories. \textbf{Single-document questions} (L01--L04) require locating and extracting information from a single document, ranging from direct surface-level extraction to complex conditional extraction. \textbf{Aggregation questions} (L05--L10) require synthesizing information across multiple documents, including counting, summation, comparison, enumeration, multi-hop reasoning, and temporal queries. These are particularly challenging because they require the model to identify all relevant documents, extract the relevant facts from each, and perform the required computation correctly. \textbf{Hallucination probe questions} (L11--L12) are designed to detect fabrication: L11 questions ask about entities that do not appear anywhere in the corpus (any specific answer is definitively fabricated), while L12 questions ask about optional fields that are absent from specific documents.

\subsection{Fidelity Metrics}

We report four metrics corresponding to distinct failure modes:

\begin{itemize}
    \item \textbf{Faithfulness} (L01--L04 + L11--L12): The broadest metric, measuring accuracy on all questions where the model had sufficient information to answer correctly. This encompasses both grounding failures and fabrication, aligning with the colloquial enterprise definition of ``hallucination'' as any confidently wrong answer when correct information was available.
    \item \textbf{Grounding} (L01--L04): Accuracy on single-document questions only, isolating retrieval and comprehension errors from fabrication.
    \item \textbf{Fabrication} (L11--L12): Error rate on hallucination probe questions. Because L11 questions ask about non-existent entities, any specific answer is definitively fabricated --- there is no ambiguity about the failure mode. This provides the cleanest signal for measuring a model's tendency to invent information.
    \item \textbf{Aggregation} (L05--L10): Reported separately as a capability metric rather than a hallucination metric. Aggregation errors conflate multiple failure modes (incomplete document retrieval, computation errors, working memory limitations) that are distinct from hallucination in the traditional sense.
\end{itemize}

Additionally, we track \textbf{truncation rate} --- the percentage of responses where the model's output was cut off by the maximum token limit before completing its answer, indicating coherence loss or infinite generation.

\section{Experimental Design}\label{sec:design}

We applied the RIKER methodology at significantly expanded scale compared to the original study. Table~\ref{tab:experimental-scale} summarizes the evaluation scope.

\begin{table}[H]
\centering
\caption{Experimental Scale}
\label{tab:experimental-scale}
\begin{tabular}{@{}lrrrrr@{}}
\toprule
\textbf{Context} & \textbf{Models} & \textbf{Total Runs} & \textbf{Input Tokens} & \textbf{Output Tokens} & \textbf{Total Tokens} \\
\midrule
32K   & 35 & 1,856 & 17.98B & 181M  & 18.16B \\
128K  & 26 & 1,480 & 71.99B & 510M  & 72.50B \\
200K  & 11 &   928 & 80.68B & 418M  & 81.09B \\
\midrule
\textbf{Total} & \textbf{35} & \textbf{4,264} & \textbf{170.65B} & \textbf{1,109M} & \textbf{171.76B} \\
\bottomrule
\end{tabular}
\vspace{0.5em}

\noindent\scriptsize\textit{Every model was evaluated at four temperature settings (0.0, 0.4, 0.7, 1.0) with 8 runs per configuration. Token counts reflect total compute consumed across all runs, including runs with partial failures.}
\end{table}

\subsection{Models}

We evaluated 35 models spanning seven model families and a wide range of sizes, from 1B to 480B parameters. The model roster includes: DeepSeek V3 and V3.1; GLM 4.5, 4.5 Air, 4.6, and 4.7; Granite 4.0 H (Tiny, Micro, Small); Llama 3.1 (8B, 70B, 405B), Llama 3.2 (1B, 3B), Llama 3.3 70B, and Llama 4 (Scout, Maverick); MiniMax M2 and M2.1; Qwen 2.5 (7B--72B, including Coder variants), Qwen3 (4B--32B), Qwen3 MoE variants (30B-A3B, 235B-A22B, Coder 30B-A3B, Coder 480B-A35B, and Next 80B-A3B), including FP8 quantized variants where applicable.

Not all models were evaluated at all context lengths. All 35 models were tested at 32K; 26 at 128K; and 11 at 200K. Models were excluded from longer contexts when they did not natively support the required context length. Table~\ref{tab:model-matrix} shows the full model-context matrix.

\begin{table}[H]
\centering
\caption{Model--Context Length Matrix}
\label{tab:model-matrix}
\begin{tabular}{@{}lccc@{}}
\toprule
\textbf{Model} & \textbf{32K} & \textbf{128K} & \textbf{200K} \\
\midrule
DeepSeek V3 & \checkmark & \checkmark & -- \\
DeepSeek V3.1 & \checkmark & \checkmark & -- \\
\midrule
GLM 4.5 & \checkmark & \checkmark & -- \\
GLM 4.5 Air & \checkmark & \checkmark & -- \\
GLM 4.6 & \checkmark & \checkmark & \checkmark \\
GLM 4.7 & \checkmark & \checkmark & \checkmark \\
\midrule
Granite 4.0 H Tiny & \checkmark & \checkmark & -- \\
Granite 4.0 H Micro & \checkmark & \checkmark & -- \\
Granite 4.0 H Small & \checkmark & \checkmark & -- \\
\midrule
Llama 3.1 8B Instruct & \checkmark & \checkmark & -- \\
Llama 3.1 70B Instruct & \checkmark & \checkmark & -- \\
Llama 3.1 405B Instruct & \checkmark & \checkmark & -- \\
Llama 3.2 1B Instruct & \checkmark & \checkmark & -- \\
Llama 3.2 3B Instruct & \checkmark & \checkmark & -- \\
Llama 3.3 70B Instruct & \checkmark & \checkmark & -- \\
Llama 4 Maverick 17B-128E & \checkmark & \checkmark & \checkmark \\
Llama 4 Scout 17B-16E & \checkmark & \checkmark & \checkmark \\
\midrule
MiniMax M2 & \checkmark & \checkmark & -- \\
MiniMax M2.1 & \checkmark & \checkmark & -- \\
\midrule
Qwen 2.5 Coder 7B & \checkmark & -- & -- \\
Qwen 2.5 14B Instruct & \checkmark & -- & -- \\
Qwen 2.5 Coder 14B & \checkmark & -- & -- \\
Qwen 2.5 32B Instruct & \checkmark & -- & -- \\
Qwen 2.5 72B Instruct & \checkmark & -- & -- \\
Qwen3 4B & \checkmark & -- & -- \\
Qwen3 4B Instruct 2507 & \checkmark & \checkmark & \checkmark \\
Qwen3 8B & \checkmark & -- & -- \\
Qwen3 14B & \checkmark & -- & -- \\
Qwen3 30B-A3B & \checkmark & \checkmark & \checkmark \\
Qwen3 32B & \checkmark & -- & -- \\
Qwen3 235B-A22B & \checkmark & \checkmark & \checkmark \\
Qwen3 235B-A22B FP8 & \checkmark & \checkmark & \checkmark \\
Qwen3 Coder 30B-A3B & \checkmark & \checkmark & \checkmark \\
Qwen3 Coder 480B-A35B & \checkmark & \checkmark & \checkmark \\
Qwen3 Next 80B-A3B & \checkmark & \checkmark & \checkmark \\
\midrule
\textbf{Total} & \textbf{35} & \textbf{26} & \textbf{11} \\
\bottomrule
\end{tabular}
\end{table}

\subsection{Hardware Platforms}

Experiments were conducted across three hardware platforms to validate cross-hardware consistency:

\begin{itemize}
    \item \textbf{NVIDIA H200} --- Two servers with 8$\times$H200 GPUs (1,928 runs; 77.49B tokens)
    \item \textbf{AMD MI300X} --- Four servers with 8$\times$MI300X GPUs (2,088 runs; 85.94B tokens)
    \item \textbf{Intel Gaudi3} --- One server with 8$\times$Gaudi3 accelerators (248 runs; 8.33B tokens)
\end{itemize}

Models were served using vLLM across all platforms. Where a model was tested on multiple platforms, identical prompts, temperature settings, and scoring procedures were used to enable direct cross-hardware comparison.

\subsection{Temperature and Repetition}

Each model-context-hardware configuration was evaluated at four temperature settings: 0.0, 0.4, 0.7, and 1.0. Each configuration was run 8 times to enable statistical analysis, including computation of means, standard deviations, coefficients of variation, and 95\% confidence intervals. This multi-run design is essential because non-zero temperatures introduce stochastic variation in model outputs, and single-run evaluations cannot distinguish genuine capability differences from sampling noise.

\section{Results}\label{sec:results}

\subsection{Cross-Hardware Consistency}\label{sec:hardware}

Before analyzing hallucination rates, we first establish that hardware platform does not meaningfully affect model behavior. This is important both for the validity of pooling results across platforms in subsequent analysis and as a practical finding for deployment decisions.

Table~\ref{tab:cross-hardware} compares overall accuracy and fabrication rates for models evaluated on both NVIDIA H200 and AMD MI300X platforms at 32K context with temperature 0.0. Where a model was run on two H200 servers, we report the mean.

\begin{table}[H]
\centering
\caption{Cross-Platform Comparison at 32K Context, Temperature 0.0}
\label{tab:cross-hardware}
\begin{tabular}{@{}lrrrrrrr@{}}
\toprule
& \multicolumn{3}{c}{\textbf{Overall Accuracy (\%)}} & \multicolumn{3}{c}{\textbf{Fabrication Rate (\%)}} \\
\cmidrule(lr){2-4} \cmidrule(lr){5-7}
\textbf{Model} & \textbf{H200} & \textbf{MI300X} & \textbf{$\Delta$} & \textbf{H200} & \textbf{MI300X} & \textbf{$\Delta$} \\
\midrule
GLM 4.5              & 97.01 & 97.40 & 0.39 & 1.59 & 1.88 & 0.30 \\
Qwen3 Coder 480B-A35B & 89.80 & 90.54 & 0.75 & 14.93 & 13.00 & 1.93 \\
Qwen3 235B-A22B       & 89.26 & 90.20 & 0.94 & 22.82 & 20.24 & 2.58 \\
Qwen3 235B-A22B FP8 & 87.97 & 88.21 & 0.24 & 25.25 & 24.31 & 0.94 \\
GLM 4.6               & 92.81 & 93.26 & 0.46 & 8.68 & 7.04 & 1.64 \\
GLM 4.5 Air            & 92.15 & 91.24 & 0.91 & 5.70 & 3.37 & 2.33 \\
DeepSeek V3            & 79.98 & 79.64 & 0.34 & 28.87 & 28.57 & 0.30 \\
\bottomrule
\end{tabular}
\vspace{0.5em}

\noindent\scriptsize\textit{H200 values are averaged across two servers where applicable. $\Delta$ = absolute difference between platforms. All models were served using vLLM with identical prompts and scoring.}
\end{table}

Overall accuracy differences between H200 and MI300X are small, ranging from 0.24 to 0.94 percentage points (mean $\Delta$ = 0.58). For context, these cross-hardware deltas are consistently smaller than the variation induced by changing temperature on the \textit{same} hardware. Table~\ref{tab:hw-vs-temp} illustrates this: for every model examined, the temperature-induced spread exceeds the hardware-induced difference by a factor of 1.3$\times$ to 9.4$\times$.

\begin{table}[H]
\centering
\caption{Hardware Delta vs.\ Temperature Spread at 32K Context}
\label{tab:hw-vs-temp}
\begin{tabular}{@{}lrrr@{}}
\toprule
\textbf{Model} & \textbf{Hardware $\Delta$} & \textbf{Temp Spread} & \textbf{Ratio} \\
\midrule
GLM 4.5              & 0.39 & 1.28 & 3.3$\times$ \\
Qwen3 Coder 480B-A35B & 0.75 & 2.59 & 3.5$\times$ \\
Qwen3 235B-A22B       & 0.94 & 1.25 & 1.3$\times$ \\
GLM 4.6               & 0.46 & 2.32 & 5.0$\times$ \\
GLM 4.5 Air            & 0.91 & 3.62 & 4.0$\times$ \\
DeepSeek V3            & 0.34 & 3.20 & 9.4$\times$ \\
\bottomrule
\end{tabular}
\vspace{0.5em}

\noindent\scriptsize\textit{Hardware $\Delta$ = absolute difference in overall accuracy between H200 and MI300X at temperature 0.0. Temperature Spread = range of overall accuracy across four temperature settings (0.0, 0.4, 0.7, 1.0) on a single platform. Ratio = Temp Spread / Hardware $\Delta$.}
\end{table}

Fabrication rates show the same pattern. While fabrication deltas are slightly larger than overall accuracy deltas (up to 2.58 percentage points for Qwen3 235B-A22B), these differences are comparable in magnitude to the variation observed between two servers of the \textit{same} hardware platform. For instance, Qwen3 235B-A22B's fabrication rate on the two H200 servers was 24.01\% and 21.63\% --- a within-platform spread of 2.38 points, which is nearly as large as the cross-platform delta of 2.58 points. This indicates that the observed variation reflects run-to-run noise rather than a systematic hardware effect.

The Intel Gaudi3 platform, while tested on fewer models, confirms the pattern. Table~\ref{tab:three-platform} compares models evaluated on Gaudi3 alongside MI300X (and H200 where available) at temperature 1.0. Across a wide performance range --- from Qwen3 235B-A22B at 89\% to Llama 3.2 3B at 34\% --- the maximum cross-platform delta averages 0.82 points. The largest deltas occur for smaller models (e.g., Llama 3.2 3B: 1.45 points), consistent with their higher run-to-run variance at temperature 1.0 rather than any systematic hardware effect.

\begin{table}[H]
\centering
\caption{Three-Platform Comparison at 32K Context, Temperature 1.0}
\label{tab:three-platform}
\begin{tabular}{@{}lrrrr@{}}
\toprule
\textbf{Model} & \textbf{H200} & \textbf{MI300X} & \textbf{Gaudi3} & \textbf{Max $\Delta$} \\
\midrule
Qwen3 235B-A22B       & 88.80 & 88.95 & 88.68 & 0.27 \\
Qwen2.5 72B Instruct  & ---   & 80.29 & 80.63 & 0.34 \\
Qwen3 30B-A3B         & ---   & 76.85 & 77.09 & 0.24 \\
Qwen3 Coder 30B-A3B   & ---   & 77.19 & 76.38 & 0.81 \\
Qwen2.5 Coder 7B      & ---   & 50.47 & 50.27 & 0.20 \\
Llama 3.1 8B Instruct & ---   & 48.01 & 48.92 & 0.91 \\
Llama 3.2 3B Instruct & ---   & 33.89 & 35.34 & 1.45 \\
\bottomrule
\end{tabular}
\vspace{0.5em}

\noindent\scriptsize\textit{H200 values averaged across servers where applicable. ``---'' indicates the model was not tested on that platform at this temperature. Models selected to span the full performance range; remaining Gaudi3 models show comparable cross-platform agreement.}
\end{table}

These findings hold at longer context lengths. At 128K context, GLM 4.5 scored 87.32\% (H200) vs.\ 86.71\% (MI300X), a delta of 0.61 points. At 200K context, Qwen3 Coder 480B-A35B scored 71.08\% (H200) vs.\ 71.39\% (MI300X), a delta of 0.31 points. Grounding and aggregation metrics show similarly narrow cross-hardware gaps.

One notable exception is DeepSeek V3.1 at 32K context, which showed a 2--3 point overall accuracy advantage on H200 over MI300X, with a larger divergence in fabrication rates: MI300X produced an identical 14.38\% fabrication rate across temperatures 0.0, 0.4, and 0.7 --- an anomalous lack of temperature sensitivity --- while H200 showed the expected gradual increase from 6.35\% to 8.43\%. This artifact appears to be 32K-specific; at 128K context, MI300X fabrication rates vary normally with temperature (7.71\% at $t$=0.0 to 9.79\% at $t$=1.0), and the best single result across either platform (90.45\% overall accuracy) was achieved on MI300X at 128K. DeepSeek V3.1 was not tested at 200K context. We suspect the 32K artifact reflects differences in vLLM versions between platforms: the MI300X servers used a custom ROCm-based vLLM image built on October 14, 2025, while the H200 servers used \texttt{vllm/vllm-openai:latest} pulled in December 2025 --- approximately two months newer. Anecdotally and outside of the experiments done for this study, the researchers have also previously encountered DeepSeek V3.1 measurements that went the opposite way, seemingly scoring lower on H200 compared to MI300X, but also not replicable once we controlled for vLLM versions.

Importantly, this anomaly is specific to a single model at a single context length; no other model exhibited comparable cross-platform divergence.

\textbf{Conclusion.} Hardware platform does not meaningfully affect model evaluation outcomes. Cross-hardware differences are smaller than temperature-induced variation and far smaller than inter-model differences. For the remainder of this paper, we pool results across hardware platforms, reporting the mean across all available runs for each model-temperature-context configuration. Readers should note that DeepSeek V3.1 at 32K on MI300X exhibits a serving artifact that inflates its fabrication rate; its H200 results better represent the model's actual capability at that context length.

\subsection{Overall Performance at 32K Context}\label{sec:overall-32k}

Table~\ref{tab:32k-leaderboard} presents the full results for all models at 32K context, ranked by overall accuracy at each model's best-performing temperature. For each model, we select the hardware platform and temperature that produced the highest overall accuracy, consistent with our finding in Section~\ref{sec:hardware} that hardware does not meaningfully affect results.

\begin{longtable}{@{}rlrrrrr@{}}
\caption{Model Performance at 32K Context (Best Temperature per Model)} \label{tab:32k-leaderboard} \\
\toprule
\textbf{\#} & \textbf{Model} & \textbf{Overall} & \textbf{Faith.} & \textbf{Ground.} & \textbf{Fab.$\downarrow$} & \textbf{Agg.} \\
\midrule
\endfirsthead
\multicolumn{7}{c}{\textit{Table~\ref{tab:32k-leaderboard} continued}} \\
\toprule
\textbf{\#} & \textbf{Model} & \textbf{Overall} & \textbf{Faith.} & \textbf{Ground.} & \textbf{Fab.$\downarrow$} & \textbf{Agg.} \\
\midrule
\endhead
\bottomrule
\endfoot
1  & GLM 4.5                  & 97.40 & 97.92 & 97.72 &  1.88 & 96.32 \\
2  & MiniMax M2.1             & 95.96 & 96.48 & 98.21 &  5.26 & 94.85 \\
3  & DeepSeek V3.1            & 95.49 & 95.44 & 97.22 &  6.35 & 95.59 \\
4  & MiniMax M2               & 95.25 & 94.99 & 96.63 &  6.65 & 95.80 \\
5  & Qwen3 Next 80B-A3B       & 93.87 & 93.70 & 94.44 &  7.04 & 94.22 \\
6  & GLM 4.6                  & 93.26 & 93.50 & 94.05 &  7.04 & 92.75 \\
7  & GLM 4.5 Air              & 92.45 & 92.66 & 91.07 &  5.75 & 92.02 \\
8  & Qwen3 Coder 480B-A35B    & 91.21 & 90.33 & 92.86 & 12.20 & 93.07 \\
9  & Qwen3 235B-A22B          & 90.20 & 88.00 & 96.23 & 20.24 & 94.85 \\
10 & Qwen3 235B-A22B FP8    & 88.51 & 85.86 & 95.34 & 23.61 & 94.12 \\
11 & Llama 4 Maverick         & 86.52 & 84.03 & 96.73 & 28.67 & 91.81 \\
12 & GLM 4.7                  & 86.42 & 88.89 & 95.34 & 17.56 & 81.20 \\
13 & Qwen 2.5 72B             & 84.77 & 86.51 & 94.44 & 21.43 & 81.09 \\
14 & Llama 3.1 405B           & 84.75 & 84.29 & 95.08 & 26.51 & 85.71 \\
15 & Qwen3 32B                & 83.42 & 84.77 & 90.87 & 21.33 & 80.57 \\
16 & Qwen 2.5 32B             & 80.66 & 83.09 & 88.29 & 22.12 & 75.53 \\
17 & DeepSeek V3              & 79.98 & 79.66 & 88.19 & 28.87 & 80.67 \\
18 & Qwen3 4B (2507)          & 79.38 & 81.15 & 88.19 & 25.89 & 75.63 \\
19 & Qwen3 30B-A3B            & 77.97 & 76.04 & 88.19 & 36.11 & 82.04 \\
20 & Qwen3 Coder 30B-A3B      & 77.19 & 74.75 & 89.78 & 40.28 & 82.35 \\
21 & Qwen 2.5 14B             & 74.36 & 78.57 & 82.94 & 25.79 & 65.44 \\
22 & Qwen 2.5 Coder 14B       & 70.96 & 80.56 & 87.10 & 25.99 & 50.63 \\
23 & Llama 3.1 70B            & 69.76 & 68.45 & 89.95 & 53.04 & 72.55 \\
24 & Qwen3 14B                & 68.49 & 80.21 & 82.84 & 22.42 & 43.70 \\
25 & Llama 4 Scout            & 68.26 & 68.90 & 82.24 & 44.44 & 66.91 \\
26 & Llama 3.3 70B            & 66.64 & 69.39 & 80.46 & 41.67 & 60.82 \\
27 & Qwen3 8B                 & 65.97 & 77.13 & 81.75 & 27.48 & 42.33 \\
28 & Qwen3 4B                 & 55.33 & 61.36 & 80.16 & 57.44 & 42.54 \\
29 & Granite 4.0 H Small      & 54.35 & 62.35 & 67.76 & 43.06 & 37.39 \\
30 & Llama 3.1 8B             & 54.11 & 63.49 & 66.96 & 39.98 & 34.24 \\
31 & Qwen 2.5 Coder 7B        & 53.27 & 60.81 & 67.86 & 46.23 & 37.29 \\
32 & Llama 3.2 3B             & 42.67 & 54.25 & 50.79 & 42.29 & 18.13 \\
33 & Granite 4.0 H Micro      & 38.37 & 46.28 & 46.33 & 53.77 & 21.64 \\
34 & Granite 4.0 H Tiny       & 32.77 & 33.96 & 47.39 & 79.48 & 30.25 \\
35 & Llama 3.2 1B             & 24.80 & 33.73 & 14.95 & 47.49 &  5.88 \\
\end{longtable}

\noindent\scriptsize\textit{All values are percentages. Overall, Faithfulness, Grounding, and Aggregation: higher is better. Fabrication ($\downarrow$): lower is better (measures the rate at which models fabricate answers to questions about non-existent entities). Each model is shown at its best-performing temperature and hardware platform. Truncation rates are omitted for space; most models show $<$1\% truncation at 32K, with notable exceptions discussed in Section~\ref{sec:temp-coherence}.}

\normalsize

Several patterns emerge from Table~\ref{tab:32k-leaderboard}:

\textbf{Model size is necessary but not sufficient.} While the top performers are generally large models, size alone does not determine ranking. Qwen3 Next 80B-A3B (a mixture-of-experts model with only 3B active parameters) ranks 5th at 93.87\%, outperforming much larger dense models including Llama 3.1 405B (84.75\%) and Qwen 2.5 72B (84.77\%). Conversely, Llama 3.1 70B (69.76\%) substantially underperforms several models a fraction of its size.

\textbf{Grounding and fabrication are distinct capabilities.} This is visible even in the overall rankings. Llama 3.1 70B achieves 89.95\% grounding (correctly extracting facts from documents) yet fabricates answers 53.04\% of the time --- meaning it is good at finding facts that exist but equally willing to invent facts that do not. This pattern recurs across multiple models and is analyzed in detail in Section~\ref{sec:grounding-vs-fab}.

\textbf{Aggregation is the hardest task category.} For most models, aggregation accuracy is the lowest of the three positive metrics. The gap between grounding and aggregation exceeds 20 percentage points for many mid-tier models (e.g., Qwen3 14B: 82.84\% grounding vs.\ 43.70\% aggregation), confirming the original RIKER finding that cross-document synthesis remains substantially harder than single-document extraction.

\subsection{Context Length Degradation}\label{sec:context-degradation}

A central question for practical deployment is whether models maintain their fidelity as context length increases. Of our 35 models, 26 were tested at both 32K and 128K tokens, and 11 at all three context lengths (32K, 128K, and 200K). The results are unambiguous: \textbf{every model degrades}, though the magnitude varies dramatically.

\subsubsection{128K Context Performance}

Table~\ref{tab:128k-degradation} presents the 26 models tested at both 32K and 128K context lengths, ranked by 128K overall accuracy. Each model is shown at its best-performing temperature and hardware platform at each context length.

\begin{longtable}{rlrrrrr}
\caption{Performance Degradation from 32K to 128K Context (26 Models)}\label{tab:128k-degradation} \\
\toprule
\textbf{Rank} & \textbf{Model} & \textbf{32K\%} & \textbf{128K\%} & \textbf{$\Delta$} & \textbf{Fab 32K} & \textbf{Fab 128K} \\
\midrule
\endfirsthead
\multicolumn{7}{c}{\tablename\ \thetable\ (continued)} \\
\toprule
\textbf{Rank} & \textbf{Model} & \textbf{32K\%} & \textbf{128K\%} & \textbf{$\Delta$} & \textbf{Fab 32K} & \textbf{Fab 128K} \\
\midrule
\endhead
\bottomrule
\endfoot
 1 & DeepSeek V3.1          & 95.49 & 90.45 & $-$5.04  &  6.35 &  7.71 \\
 2 & MiniMax M2             & 95.25 & 89.27 & $-$5.98  &  6.65 &  8.47 \\
 3 & Qwen3 Next 80B-A3B    & 93.87 & 87.85 & $-$6.02  &  7.04 &  7.99 \\
 4 & GLM 4.5                & 97.40 & 87.43 & $-$9.97  &  1.29 &  3.19 \\
 5 & GLM 4.6                & 93.26 & 85.81 & $-$7.45  &  7.04 & 13.75 \\
 6 & MiniMax M2.1           & 95.96 & 85.59 & $-$10.37 &  5.06 &  9.72 \\
 7 & GLM 4.7                & 86.42 & 83.20 & $-$3.22  & 17.56 & 10.49 \\
 8 & Qwen3 Coder 480B-A35B & 91.21 & 80.04 & $-$11.17 & 12.20 & 16.88 \\
 9 & Qwen3 235B-A22B       & 90.20 & 79.90 & $-$10.30 & 20.24 & 22.08 \\
10 & Qwen3 235B-A22B FP8   & 88.51 & 79.79 & $-$8.72  & 23.61 & 23.54 \\
11 & GLM 4.5 Air            & 92.45 & 72.00 & $-$20.45 &  3.37 & 12.43 \\
12 & Qwen3 30B-A3B 2507    & 77.97 & 69.91 & $-$8.06  & 36.11 & 35.07 \\
13 & DeepSeek V3             & 79.98 & 68.56 & $-$11.42 & 28.57 & 30.49 \\
14 & Llama 4 Maverick       & 86.52 & 63.90 & $-$22.62 & 28.67 & 38.82 \\
15 & Qwen3 Coder 30B-A3B   & 77.19 & 63.38 & $-$13.81 & 40.28 & 43.82 \\
16 & Qwen3 4B 2507          & 79.38 & 59.57 & $-$19.81 & 25.89 & 31.11 \\
17 & Llama 3.1 405B         & 84.75 & 58.29 & $-$26.46 & 26.51 & 30.62 \\
18 & Llama 4 Scout          & 68.26 & 49.91 & $-$18.35 & 44.44 & 52.22 \\
19 & Llama 3.1 8B           & 54.11 & 44.15 & $-$9.96  & 39.98 & 32.15 \\
20 & Granite 4.0 H Small    & 54.35 & 43.91 & $-$10.44 & 43.06 & 34.44 \\
21 & Llama 3.1 70B          & 69.76 & 42.08 & $-$27.68 & 53.04 & 56.67 \\
22 & Llama 3.3 70B          & 66.64 & 38.36 & $-$28.28 & 41.67 & 53.89 \\
23 & Llama 3.2 3B           & 42.67 & 31.02 & $-$11.65 & 42.29 & 56.74 \\
24 & Granite 4.0 H Micro    & 38.37 & 30.88 & $-$7.49  & 53.77 & 49.79 \\
25 & Llama 3.2 1B           & 24.80 & 28.60 & $+$3.80  & 47.49 & 28.33 \\
26 & Granite 4.0 H Tiny     & 32.77 & 21.05 & $-$11.72 & 79.48 & 86.46 \\
\end{longtable}

\noindent\scriptsize\textit{All values are percentages. Overall accuracy (higher is better); Fabrication ($\downarrow$): lower is better. $\Delta$ = 128K score minus 32K score. Each model shown at its best-performing temperature and platform at each context length independently. Rank is by 128K overall accuracy.}

\normalsize

Across the 26 models, the median degradation from 32K to 128K is 10.4 percentage points, with a mean of 12.4~pp. However, the range spans from a 3.2~pp drop (GLM~4.7) to a 28.3~pp collapse (Llama~3.3~70B). Several patterns emerge:

\textbf{The Llama family degrades disproportionately.} Llama~3.1~70B ($-$27.7~pp), Llama~3.3~70B ($-$28.3~pp), and Llama~3.1~405B ($-$26.5~pp) all lose more than 25 percentage points moving to 128K context. The 405B model --- ranked 12th at 32K --- falls to 17th at 128K, behind several models a fraction of its size. Llama~4 Maverick, despite being a newer architecture, still drops 22.6~pp.

\textbf{Fabrication increases drive much of the degradation.} For most models, fabrication rates rise as context grows: Llama~3.3~70B goes from 41.67\% to 53.89\%, GLM~4.5~Air from 3.37\% to 12.43\%, and Llama~4~Scout from 44.44\% to 52.22\%. With more context, models become more willing to fabricate answers about entities that do not exist in the documents.

\textbf{GLM 4.5 Air suffers a surprising drop.} Despite ranking 7th at 32K (92.45\%), it falls to 11th at 128K (72.00\%), losing 20.45~pp --- substantially more than its GLM siblings GLM~4.5 ($-$9.97~pp) and GLM~4.6 ($-$7.45~pp).

\textbf{One model shows no degradation.} Llama~3.2~1B is the sole exception, improving by 3.8~pp from 32K to 128K. However, at absolute scores of 24.80\% and 28.60\%, this model is performing so poorly at both context lengths that the change likely reflects noise rather than genuine improvement.

\subsubsection{200K Context Performance}

Table~\ref{tab:200k-degradation} extends the analysis to the 11 models tested at all three context lengths.

\begin{longtable}{rlrrrrr}
\caption{Performance Across All Three Context Lengths (11 Models)}\label{tab:200k-degradation} \\
\toprule
\textbf{Rank} & \textbf{Model} & \textbf{32K\%} & \textbf{128K\%} & \textbf{200K\%} & \textbf{$\Delta_{32 \rightarrow 200}$} & \textbf{Fab 200K} \\
\midrule
\endfirsthead
\multicolumn{7}{c}{\tablename\ \thetable\ (continued)} \\
\toprule
\textbf{Rank} & \textbf{Model} & \textbf{32K\%} & \textbf{128K\%} & \textbf{200K\%} & \textbf{$\Delta_{32 \rightarrow 200}$} & \textbf{Fab 200K} \\
\midrule
\endhead
\bottomrule
\endfoot
 1 & Qwen3 Next 80B-A3B    & 93.87 & 87.85 & 82.68 & $-$11.19 & 11.21 \\
 2 & Qwen3 Coder 480B-A35B & 91.21 & 80.04 & 71.67 & $-$19.54 & 27.58 \\
 3 & Qwen3 235B-A22B       & 90.20 & 79.90 & 69.93 & $-$20.27 & 35.68 \\
 4 & Qwen3 235B-A22B FP8   & 88.51 & 79.79 & 69.52 & $-$18.99 & 34.66 \\
 5 & Qwen3 30B-A3B 2507    & 77.97 & 69.91 & 65.45 & $-$12.52 & 41.04 \\
 6 & Llama 4 Maverick       & 86.52 & 63.90 & 61.56 & $-$24.96 & 43.29 \\
 7 & GLM 4.7                & 86.42 & 83.20 & 57.11 & $-$29.31 & 46.73 \\
 8 & Qwen3 Coder 30B-A3B   & 77.19 & 63.38 & 56.40 & $-$20.79 & 50.27 \\
 9 & Qwen3 4B 2507          & 79.38 & 59.57 & 51.53 & $-$27.85 & 26.02 \\
10 & Llama 4 Scout          & 68.26 & 49.91 & 46.13 & $-$22.13 & 52.09 \\
11 & GLM 4.6                & 93.26 & 85.81 & 37.65 & $-$55.61 & 71.62 \\
\end{longtable}

\noindent\scriptsize\textit{All values are percentages. Rank is by 200K overall accuracy. $\Delta_{32 \rightarrow 200}$ = total degradation from 32K to 200K. Fab 200K = fabrication rate at 200K context (lower is better).}

\normalsize

The mean degradation from 32K to 200K across these 11 models is 23.9 percentage points --- more than double the 32K-to-128K median. Two qualitatively different behaviors emerge:

\textbf{Graceful degradation.} Qwen3 Next 80B-A3B stands out as the most context-resilient model, losing only 11.19~pp across the full range and maintaining a fabrication rate of just 11.21\% even at 200K. Similarly, Qwen3 30B-A3B~2507 loses only 12.52~pp. Both are mixture-of-experts architectures with small active parameter counts, suggesting that MoE routing may confer some advantage in maintaining fidelity over long contexts.

\textbf{Catastrophic collapse.} GLM~4.6 presents the most dramatic failure: despite ranking 6th at 32K (93.26\%) and 5th at 128K (85.81\%), it collapses to last place at 200K (37.65\%). The mechanism is clear from the fabrication rate, which explodes from 7.04\% at 32K to 13.75\% at 128K to 71.62\% at 200K. At 200K context, GLM~4.6 fabricates answers to nearly three-quarters of questions about non-existent entities. GLM~4.7 follows a similar trajectory, though less extreme: it was the most resilient model at 128K ($-$3.22~pp) but then drops sharply at 200K ($-$29.31~pp total, fabrication rising to 46.73\%).

\textbf{Aggregation degrades fastest.} Across all models, aggregation accuracy drops more steeply than other metrics as context grows. For example, Qwen3 Coder 480B-A35B maintains reasonable grounding at 200K (81.71\%) but its aggregation falls from 93.07\% at 32K to 60.70\% at 200K --- a 32~pp drop. This pattern is consistent: cross-document synthesis becomes disproportionately harder as the number of documents grows with context length.

\textbf{Context length claims frequently exceed usable capacity.} Many models that officially support 128K or 200K tokens show degradation so severe that their effective capacity is far lower. GLM~4.6 at 200K (37.65\%) performs worse than Llama~3.2~3B at 32K (42.67\%), a model with a fraction of its parameters and a fraction of the context. Llama~3.1~405B at 128K (58.29\%) falls below Qwen3~4B at 32K (79.38\%). These results suggest that \textit{advertised} context length is a poor proxy for \textit{usable} context length in document Q\&A tasks.

\subsection{How Much Do They Hallucinate?}\label{sec:fabrication}

The preceding sections reported fabrication rates at each model's best \textit{overall accuracy} temperature. But fabrication can sometimes be reduced further by choosing a temperature that minimizes fabrication specifically, even if that temperature is not optimal for overall performance. Table~\ref{tab:fabrication-floor} presents each model's \textbf{best-case fabrication rate} --- the lowest fabrication achievable at any temperature and hardware platform. This represents the floor: even when given every advantage, how often does each model still hallucinate?

\begin{longtable}{rlrrrr}
\caption{Best-Case Fabrication Rates Across Context Lengths ($\downarrow$ lower is better)}\label{tab:fabrication-floor} \\
\toprule
\textbf{Rank} & \textbf{Model} & \textbf{32K} & \textbf{128K} & \textbf{200K} & \textbf{$\Delta_{32 \rightarrow \text{max}}$} \\
\midrule
\endfirsthead
\multicolumn{6}{c}{\tablename\ \thetable\ (continued)} \\
\toprule
\textbf{Rank} & \textbf{Model} & \textbf{32K} & \textbf{128K} & \textbf{200K} & \textbf{$\Delta_{32 \rightarrow \text{max}}$} \\
\midrule
\endhead
\bottomrule
\endfoot
 1 & GLM 4.5                &  1.19 &  3.19 &   --- & $+$2.00 \\
 2 & GLM 4.5 Air            &  3.37 & 12.43 &   --- & $+$9.06 \\
 3 & MiniMax M2.1           &  5.06 &  9.72 &   --- & $+$4.66 \\
 4 & DeepSeek V3.1          &  6.35 &  7.36 &   --- & $+$1.01 \\
 5 & MiniMax M2             &  6.55 &  7.22 &   --- & $+$0.67 \\
 6 & GLM 4.6                &  7.04 & 13.75 & 69.53 & $+$62.49 \\
 7 & Qwen3 Next 80B-A3B    &  7.04 &  7.99 & 10.25 & $+$3.21 \\
 8 & Qwen3 Coder 480B-A35B & 12.20 & 16.88 & 27.52 & $+$15.32 \\
 9 & GLM 4.7                & 16.27 & 10.42 & 46.73 & $+$30.46 \\
10 & Qwen3 235B-A22B       & 19.84 & 22.08 & 33.15 & $+$13.31 \\
11 & Qwen3 14B             & 20.63 &   --- &   --- &     --- \\
12 & Qwen3 32B             & 21.03 &   --- &   --- &     --- \\
13 & Qwen 2.5 72B          & 21.43 &   --- &   --- &     --- \\
14 & Qwen 2.5 32B          & 21.43 &   --- &   --- &     --- \\
15 & Qwen3 235B-A22B FP8   & 23.31 & 21.81 & 32.94 & $+$9.63 \\
16 & Qwen3 4B 2507         & 25.30 & 29.79 & 26.02 & $+$4.49 \\
17 & Qwen 2.5 14B          & 25.60 &   --- &   --- &     --- \\
18 & Qwen 2.5 Coder 14B    & 25.79 &   --- &   --- &     --- \\
19 & DeepSeek V3            & 26.49 & 30.49 &   --- & $+$4.00 \\
20 & Llama 3.1 405B        & 26.51 & 30.62 &   --- & $+$4.11 \\
21 & Qwen3 8B              & 27.48 &   --- &   --- &     --- \\
22 & Llama 4 Maverick      & 28.08 & 38.82 & 43.29 & $+$15.21 \\
23 & Qwen3 30B-A3B 2507    & 36.11 & 35.07 & 41.04 & $+$4.93 \\
24 & Llama 3.1 8B          & 39.98 & 32.15 &   --- & $-$7.83 \\
25 & Qwen3 Coder 30B-A3B   & 40.18 & 43.82 & 47.21 & $+$7.03 \\
26 & Llama 3.3 70B         & 41.67 & 53.89 &   --- & $+$12.22 \\
27 & Llama 3.2 3B          & 42.29 & 56.74 &   --- & $+$14.45 \\
28 & Granite 4.0 H Small   & 43.06 & 34.44 &   --- & $-$8.62 \\
29 & Llama 4 Scout         & 44.44 & 52.22 & 52.09 & $+$7.65 \\
30 & Qwen 2.5 Coder 7B    & 45.83 &   --- &   --- &     --- \\
31 & Llama 3.2 1B          & 47.35 & 28.33 &   --- & $-$19.02 \\
32 & Llama 3.1 70B         & 49.50 & 56.67 &   --- & $+$7.17 \\
33 & Granite 4.0 H Micro   & 52.58 & 49.24 &   --- & $-$3.34 \\
34 & Qwen3 4B              & 57.44 &   --- &   --- &     --- \\
35 & Granite 4.0 H Tiny    & 78.27 & 85.76 &   --- & $+$7.49 \\
\end{longtable}

\noindent\scriptsize\textit{All values are fabrication rate percentages (lower is better). Each model is shown at its best temperature and platform for minimizing fabrication at each context length independently. $\Delta_{32 \rightarrow \text{max}}$ = change from 32K to the longest context tested. ``---'' indicates the model was not tested at that context length.}

\normalsize

The results are sobering. At 32K context --- the shortest and easiest setting --- only 7 of 35 models (20\%) achieve a fabrication rate below 10\%, and only 2 models (GLM~4.5 at 1.19\% and GLM~4.5~Air at 3.37\%) stay below 5\%. At 128K, only 5 of 26 models remain under 10\%. At 200K, \textbf{no model achieves a fabrication rate below 10\%}; the best performer, Qwen3 Next 80B-A3B, still fabricates answers to 10.25\% of questions about non-existent entities.

Focusing first on 32K---the baseline context length where all 35 models were tested---several patterns emerge:

\textbf{Fabrication rates cluster into tiers.} A clear gap separates the top tier ($<$10\% fabrication: GLM~4.5, GLM~4.5~Air, MiniMax~M2/M2.1, DeepSeek~V3.1, GLM~4.6, Qwen3~Next~80B-A3B) from the mid tier (12--28\%: most Qwen variants, DeepSeek~V3, Llama~3.1~405B) and the high-fabrication tier ($>$36\%: smaller Llama models, Granite models, Qwen3~4B). Nearly half of all models (17 of 35) fabricate answers to more than 25\% of trap questions even at 32K.

\textbf{Model family matters more than model size.} At 32K, the GLM family achieves low fabrication across all variants tested, while the Llama~3.x family consistently shows high fabrication regardless of size --- Llama~3.1~405B (26.51\%) fabricates at a similar rate to Llama~3.1~8B (39.98\%), and Llama~3.1~70B is worse than both at 49.50\%. The Qwen family clusters in the 20--27\% range across multiple sizes. This suggests that fabrication resistance is largely a training-time property, not an emergent capability of scale---though as the next observation shows, even well-trained resistance can collapse at longer contexts.

Looking across context lengths, the table reveals a second critical dimension:

\textbf{Context length is the strongest driver of increased fabrication.} For most models, fabrication rises with context length --- but the magnitude varies enormously. GLM~4.6 shows the most extreme increase: from 7.04\% at 32K to 69.53\% at 200K ($+$62.49~pp), effectively transforming from a top-tier model into one that fabricates answers to the majority of trap questions. Similarly, GLM~4.7 rises from 16.27\% to 46.73\%. By contrast, Qwen3~Next~80B-A3B increases only from 7.04\% to 10.25\% ($+$3.21~pp), and MiniMax~M2 barely changes between 32K and 128K ($+$0.67~pp).

\textbf{A few models show reduced fabrication at longer contexts.} Llama~3.1~8B, Granite~4.0~H~Small, and Llama~3.2~1B all show lower fabrication rates at 128K than at 32K. However, all three simultaneously lose substantial grounding and overall accuracy, and are poor performers overall (ranks 29, 30, and 35 of 35 at 32K).

\textbf{How much do LLMs hallucinate in document Q\&A?} Under the best conditions we tested (32K context, optimal temperature, best hardware platform), the single best model still fabricates answers 1.19\% of the time. A more representative answer for the top tier is 5--7\%. For the median model, the answer is approximately 25\% --- one in four questions about non-existent entities receives a fabricated answer. And at 200K context, even the best model cannot stay below 10\%. These are \textit{best-case} numbers; at non-optimal temperatures or longer contexts, fabrication rates are substantially higher.

\subsection{Temperature Effects on Accuracy}\label{sec:temp-accuracy}

A common recommendation for deterministic LLM outputs is to set temperature to zero. Our data challenges this as a universal best practice. Across 90 model--context--hardware combinations, T=0.0 produces the highest overall accuracy in 60\% of cases --- a majority, but far from a rule. The remaining 40\% achieve their best overall accuracy at a higher temperature (T=0.4, 0.7, or 1.0).

Table~\ref{tab:temp-pref} summarizes the temperature preference breakdown across all five fidelity metrics.

\begin{table}[h]
\centering
\caption{Temperature Preference by Metric (90 Model--Context--Hardware Combinations)}\label{tab:temp-pref}
\begin{tabular}{lrr}
\toprule
\textbf{Metric} & \textbf{Higher Temp Best} & \textbf{T=0.0 Best} \\
\midrule
Overall Accuracy   & 36 (40\%) & 54 (60\%) \\
Faithfulness       & 41 (46\%) & 49 (54\%) \\
Grounding          & 48 (53\%) & 42 (47\%) \\
Fabrication$^*$    & 48 (53\%) & 42 (47\%) \\
Aggregation        & 39 (43\%) & 51 (57\%) \\
\bottomrule
\end{tabular}

\noindent\scriptsize $^*$For fabrication, ``higher temp best'' means higher temperature produces \textit{lower} (better) fabrication rates.
\end{table}

Several findings emerge:

\textbf{T=0.0 is not universally optimal for any metric.} Even for overall accuracy, where T=0.0 has its strongest showing, 40\% of combinations achieve better results at higher temperatures. For grounding and fabrication, the relationship is actually reversed: higher temperatures produce better scores in 53\% of cases. This means that the conventional wisdom of ``just set temperature to zero'' can actively \textit{increase} hallucination rates for the majority of models.

\textbf{Magnitude matters more than frequency.} While the table shows how often each temperature wins, the practical question is by how much. Most overall accuracy spreads between the best and worst temperature are small: the median spread at 32K is approximately 2~pp, and at 200K most spreads fall below 2~pp. By contrast, \textbf{aggregation} shows the largest magnitude effects when temperature does matter --- spreads of 10.61~pp (Qwen~2.5~32B at 32K) and 11.87~pp (GLM~4.6 at 128K). This makes sense: aggregation requires cross-document synthesis, where sampling diversity at higher temperatures can help or hinder the model's ability to combine information from multiple sources.

\textbf{Temperature sensitivity decreases with context length --- with exceptions.} At 32K, some models show spreads of 7--9~pp between their best and worst temperatures (e.g., Llama~3.2~3B: 8.77~pp, Llama~3.1~405B: 7.89~pp). At 200K, most spreads compress below 2~pp, suggesting that the difficulty of longer contexts overwhelms any temperature effect. The notable exception is GLM~4.6, which shows a 7.54~pp spread at 200K on MI300X --- though this model is already in a state of near-collapse at that context length (37.65\% overall accuracy).

\textbf{The practical takeaway} is that temperature tuning offers modest gains compared to model selection or context length management. The difference between the best and worst model at 32K exceeds 70 percentage points; the difference between the best and worst temperature for a given model rarely exceeds 3~pp for overall accuracy. However, for individual metrics --- especially aggregation --- temperature can matter substantially, and the optimal temperature varies by model. Blanket T=0.0 policies would have been suboptimal for roughly 40\% of the model--context configurations we tested.

\subsection{Temperature Effects on Coherence Loss}\label{sec:temp-coherence}

Section~\ref{sec:temp-accuracy} showed that T=0.0 is not universally optimal for accuracy. Here we present a stronger finding: T=0.0 dramatically \textit{increases} the rate at which models enter infinite generation loops --- a form of coherence loss where the model begins repeating or generating text indefinitely. In our experimental setup, we detect this by measuring the truncation rate: responses that exceed the maximum allowed output length are flagged as coherence failures, since they indicate the model failed to produce a coherent, terminating response.

The effect is negligible at 32K context, where most models show zero or near-zero coherence loss regardless of temperature. But at 128K and 200K, the pattern becomes dramatic and remarkably consistent: for virtually every model that exhibits coherence loss, T=0.0 produces the highest failure rate and T=1.0 produces the lowest.

Table~\ref{tab:coherence-loss} presents the models with the most significant coherence loss, showing failure rates at T=0.0 and T=1.0 alongside the ratio between them.

\begin{table}[h]
\centering
\caption{Coherence Loss Rate (\%) by Temperature (Models with Significant Effects)}\label{tab:coherence-loss}
\begin{tabular}{llrrrrl}
\toprule
\textbf{Model} & \textbf{Ctx} & \textbf{T=0.0} & \textbf{T=0.4} & \textbf{T=0.7} & \textbf{T=1.0} & \textbf{Ratio} \\
\midrule
Llama 3.1 8B        & 128K & 14.05 & 10.64 &  6.34 & 2.05 &  6.9$\times$ \\
Qwen3 4B 2507       & 200K & 13.29 & 12.41 & 11.10 & 8.44 &  1.6$\times$ \\
Granite 4.0 H Tiny  & 128K &  5.07 &  4.02 &  2.21 & 0.33 & 15.6$\times$ \\
Qwen3 Next 80B-A3B & 200K &  4.80 &  4.21 &  3.00 & 1.82 &  2.6$\times$ \\
GLM 4.6             & 200K &  4.64 &  3.79 &  2.15 & 0.14 & 34.4$\times$ \\
Llama 3.2 3B        &  32K &  3.23 &  1.65 &  1.18 & 0.20 & 16.0$\times$ \\
MiniMax M2.1        & 128K &  2.96 &  1.22 &  0.42 & 0.37 &  8.0$\times$ \\
GLM 4.7             & 200K &  2.59 &  1.57 &  0.34 & 0.05 & 48.1$\times$ \\
Llama 3.2 1B$^\dagger$  & 128K &  1.00 &  1.51 &  1.88 & 0.05 & 21.5$\times$ \\
Qwen3 235B-A22B-Instruct-2507$^\dagger$    & 128K &  0.42 &  0.27 &  0.03 & 0.08 &  5.1$\times$ \\
\bottomrule
\end{tabular}

\noindent\scriptsize\textit{Coherence loss rate = percentage of responses that entered infinite generation loops (detected via output truncation at the maximum allowed length). Ratio = T=0.0 rate divided by T=1.0 rate. Sorted by T=0.0 rate. $^\dagger$Not strictly monotonic across intermediate temperatures, though both still show substantially lower coherence loss at T=1.0 than T=0.0.}
\end{table}

\normalsize

Several aspects of this finding deserve emphasis:

\textbf{The effect is nearly universally monotonic.} As Table~\ref{tab:coherence-loss} shows, 8 of 10 affected models exhibit a perfectly monotonic decrease in coherence loss across all four temperature settings: T=0.0 $>$ T=0.4 $>$ T=0.7 $>$ T=1.0. The trend is especially clear for Llama~3.1~8B at 128K: 14.05\% $\rightarrow$ 10.64\% $\rightarrow$ 6.34\% $\rightarrow$ 2.05\%. Two models (marked $^\dagger$) show minor deviations at intermediate temperatures, but both still exhibit substantially higher coherence loss at T=0.0 than at T=1.0, and both involve rates below 2\% where small fluctuations are expected.

\textbf{Context length amplifies the effect.} At 32K, only a handful of smaller models show meaningful coherence loss, and the rates are modest (Llama~3.2~3B: 3.23\% at T=0.0). At 128K, the problem spreads to mid-tier models and the rates increase substantially. At 200K, even strong models like Qwen3~Next~80B-A3B (4.80\% at T=0.0) and GLM~4.7 (2.59\% at T=0.0) are affected --- models that show zero coherence loss at 32K.

\textbf{The accuracy--coherence tradeoff demands careful consideration at T=0.0.} Section~\ref{sec:temp-accuracy} showed that T=0.0 provides a median accuracy advantage of approximately 2~pp over other temperatures. But the coherence cost can be severe: GLM~4.6 at 200K suffers 34$\times$ more coherence loss events at T=0.0 than at T=1.0, and GLM~4.7 at the same context length shows a 48$\times$ ratio. For Llama~3.1~8B at 128K, T=0.0 causes 14\% of all queries to produce no usable response. Whether a 1--2 percentage point accuracy gain justifies a 34$\times$ increase in generation failures is a deployment-specific decision, but practitioners should be aware that the tradeoff exists and can be dramatic at longer contexts. Even seemingly negligible rates compound at enterprise scale: GLM~4.7's 0.29\% coherence loss rate at 128K with T=0.0 would produce approximately 640 failed queries per month in a deployment serving 1,000 users at 10 queries per day---several per business hour---while T=1.0 would produce none.

\textbf{The mechanism is intuitive.} At T=0.0, the model always selects the highest-probability next token. If the model enters a state where a repetitive pattern has high probability (common in long contexts where attention patterns can become self-reinforcing), greedy decoding has no mechanism to escape. Higher temperatures introduce sampling diversity that breaks these loops before they become self-sustaining. This explains both the monotonic relationship and the interaction with context length: longer contexts create more opportunities for repetitive attention patterns to form.

\subsection{Grounding and Fabrication as Distinct Capabilities}\label{sec:grounding-vs-fab}

A natural assumption is that grounding (correctly extracting facts from documents) and fabrication resistance (correctly refusing to answer questions about non-existent entities) are two sides of the same coin --- that a model good at one should be good at the other. Our data show this assumption is wrong. Grounding and fabrication are \textbf{distinct capabilities} that can vary independently.

To isolate this, Table~\ref{tab:grounding-vs-fab} pairs each model's best grounding score (at its optimal temperature for grounding) with its best fabrication score (at its optimal temperature for fabrication). If the two metrics were simply inversely related, a model with 90\% grounding should fabricate roughly 10\% of the time. The ``Gap'' column measures the deviation from this expectation: positive values indicate models that fabricate more than their grounding accuracy would predict.

\begin{table}[h]
\centering
\caption{Grounding vs.\ Fabrication: Best-Case Scores at 32K Context}\label{tab:grounding-vs-fab}
\begin{tabular}{lrrr}
\toprule
\textbf{Model} & \textbf{Grounding} & \textbf{Fabrication ($\downarrow$)} & \textbf{Gap} \\
\midrule
\multicolumn{4}{l}{\textit{Well-calibrated (Gap $<$ +5)}} \\
GLM 4.5               & 97.72 &  1.19 & $-$1.1 \\
MiniMax M2.1           & 98.21 &  5.06 & $+$3.3 \\
MiniMax M2             & 97.32 &  6.55 & $+$3.9 \\
GLM 4.6                & 94.84 &  7.04 & $+$1.9 \\
Qwen3 Next 80B-A3B    & 94.64 &  7.04 & $+$1.7 \\
DeepSeek V3.1          & 97.22 &  6.35 & $+$3.6 \\
GLM 4.5 Air            & 91.47 &  3.37 & $-$5.2 \\
\midrule
\multicolumn{4}{l}{\textit{Retrieval-strong but fabrication-prone (Gap $>$ +15)}} \\
Llama 3.1 405B         & 95.08 & 26.51 & $+$21.6 \\
Llama 4 Maverick       & 96.73 & 28.08 & $+$24.8 \\
Llama 4 Scout          & 84.23 & 44.44 & $+$28.7 \\
Qwen3 Coder 30B-A3B   & 89.78 & 40.18 & $+$30.0 \\
Qwen3 4B               & 80.16 & 57.44 & $+$37.6 \\
Llama 3.1 70B          & 90.18 & 49.50 & $+$39.7 \\
\bottomrule
\end{tabular}

\noindent\scriptsize\textit{All values are percentages at 32K context. Grounding: higher is better. Fabrication ($\downarrow$): lower is better. Gap = Grounding $+$ Fabrication $-$ 100; positive values indicate fabrication rate exceeds what grounding accuracy alone would predict. Each metric shown at its independently optimal temperature and hardware platform.}
\end{table}

\normalsize

\begin{figure}[h]
\centering
\includegraphics[width=\textwidth]{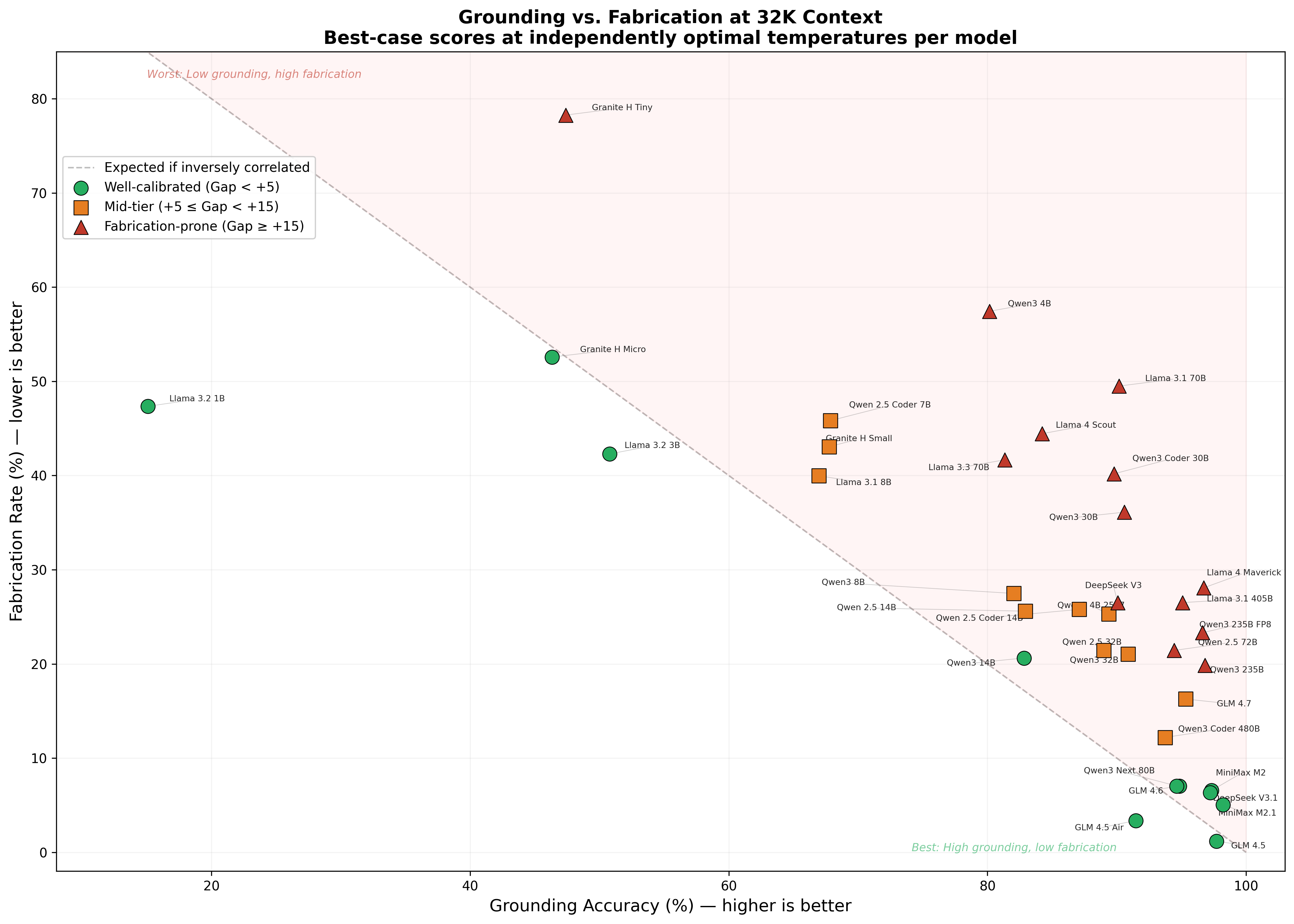}
\caption{Grounding accuracy vs.\ fabrication rate at 32K context. Each model is shown at its independently optimal temperature for each metric. The dashed line shows the expected inverse relationship; deviation above the line indicates excess fabrication. Color and shape indicate tier: well-calibrated (green circles, Gap $<$ +5), mid-tier (orange squares), and fabrication-prone (red triangles, Gap $\geq$ +15).}\label{fig:grounding-vs-fab}
\end{figure}

Figure~\ref{fig:grounding-vs-fab} visualizes this relationship across all 35 models. The dashed line represents the expected relationship if grounding and fabrication were perfectly inversely correlated; points above this line indicate models with excess fabrication relative to their grounding ability.

The contrast is stark. GLM~4.5~Air and Llama~3.1~70B achieve nearly identical grounding scores (91.47\% vs.\ 90.18\%), yet their fabrication rates differ by 46 percentage points (3.37\% vs.\ 49.50\%). Both models are equally skilled at finding facts that exist in the documents; they differ dramatically in their willingness to fabricate facts that do not. This is not an isolated case --- the Llama~3.x family consistently shows high grounding paired with high fabrication, while the GLM family and MiniMax models achieve comparable grounding with far lower fabrication.

This decoupling has important practical implications:

\textbf{Retrieval benchmarks alone are insufficient.} Many existing evaluations test only whether a model can extract correct information from documents. A model like Llama~3.1~70B would score well on such benchmarks (90\%+ grounding) while being actively dangerous in deployment --- fabricating plausible-sounding answers to nearly half of questions about information that does not exist in the source material.

\textbf{Fabrication resistance appears to be a separately trained capability.} The well-calibrated models (GLM family, MiniMax, Qwen3~Next) seem to have received specific training or alignment to resist fabrication, as their low fabrication rates cannot be explained by grounding ability alone. This suggests that reducing hallucination in document Q\&A requires targeted training interventions beyond those that improve grounding alone.

\textbf{The gap widens at longer contexts.} At 200K, even the well-calibrated GLM~4.6 sees its fabrication rate explode to 69.53\% despite maintaining 57.94\% grounding (Section~\ref{sec:context-degradation}). Fabrication resistance that holds at 32K may not transfer to longer contexts, adding another dimension to the evaluation challenge.

\section{Discussion}\label{sec:discussion}

\subsection{Practical Deployment Guidance}\label{sec:deployment}

Our results yield a clear priority ordering for practitioners deploying LLMs in document Q\&A scenarios. We present these in decreasing order of impact.

\textbf{The fabrication floor is real and non-zero.} No model in our evaluation achieves zero fabrication. The best model (GLM~4.5) still fabricates answers 1.19\% of the time at 32K under optimal conditions, and a more representative figure for top-tier models is 5--7\%. For the median model, the figure is approximately 25\%. This means that any production system using LLMs for document Q\&A must incorporate safeguards against fabricated answers --- the question is not whether the model will hallucinate, but how often and how to detect it.

\textbf{Model selection is the highest-leverage decision.} The gap between the best and worst model at 32K exceeds 70 percentage points for overall accuracy and 77 percentage points for fabrication. No amount of temperature tuning or hardware optimization can close a gap of this magnitude. Furthermore, model \textit{family} predicts fabrication resistance better than model \textit{size}: the GLM family and MiniMax models consistently achieve low fabrication across all sizes tested, while the Llama~3.x family shows high fabrication regardless of size --- Llama~3.1~405B (26.51\%) fabricates at a comparable rate to Llama~3.1~8B (39.98\%), and Llama~3.1~70B is worse than both at 49.50\%. This suggests that fabrication resistance is fundamentally a training-time property, and practitioners should prioritize models from families with demonstrated low fabrication over simply choosing the largest available model.

\textbf{Context length management is the second priority.} Every model degrades with context length, but the magnitude varies from 5~pp to over 55~pp. More critically, fabrication rates can explode at longer contexts: GLM~4.6 goes from 7\% fabrication at 32K to 70\% at 200K. For practitioners, this means that the advertised context window of a model is a poor guide to its usable capacity. Testing at the actual deployment context length is essential, as 32K benchmark results may not transfer to 128K or 200K production workloads.

\textbf{Temperature requires nuance, not dogma.} The conventional wisdom of ``always use T=0.0'' is not supported by our data. T=0.0 produces the best overall accuracy in only 60\% of cases, and higher temperatures actually reduce fabrication rates in 53\% of model--context--hardware combinations (though the magnitude is often small). More importantly, T=0.0 dramatically increases the rate of coherence loss (infinite generation loops), with ratios as high as 48$\times$ at 200K context. For enterprise deployments where query reliability matters, a moderate temperature (T=0.4 or T=0.7) often provides the best balance of accuracy, low fabrication, and low coherence loss risk.

\textbf{Hardware platform is not a meaningful variable.} Our cross-hardware analysis (Section~\ref{sec:hardware}) showed that NVIDIA~H200, AMD~MI300X, and Intel~Gaudi3 produce statistically indistinguishable results when running the same model via the same serving framework. Practitioners can select hardware based on cost, availability, and throughput considerations without concern for fidelity differences.

\subsection{Implications for Evaluation Design}\label{sec:eval-implications}

Our findings expose a significant gap in how LLM document Q\&A capabilities are commonly evaluated.

Most existing benchmarks test only \textit{grounding} --- whether a model can correctly extract information that exists in provided documents. Section~\ref{sec:grounding-vs-fab} demonstrated that this is dangerously incomplete: models like Llama~3.1~70B achieve 90\%+ grounding while fabricating answers to nearly half of questions about non-existent entities. A grounding-only evaluation would rate this model as highly capable, when in practice it poses a substantial hallucination risk.

We argue that any serious evaluation of document Q\&A fidelity must include \textbf{fabrication testing} --- questions about entities, facts, or relationships that deliberately do not exist in the provided documents. The RIKER methodology's paradigm inversion approach (generating documents from known ground truth) enables this naturally, since the ground truth defines exactly what does and does not exist. Without such testing, evaluation results can be misleading and potentially dangerous when used to guide deployment decisions.

Additionally, evaluation should be conducted at the \textit{actual deployment context length}, not merely at short contexts. A model that scores 93\% at 32K may score 38\% at 200K (as GLM~4.6 demonstrates). Single-context-length benchmarks provide false confidence about long-context performance.

\subsection{Limitations}\label{sec:limitations}

Several limitations should be considered when interpreting our results.

\textbf{Single evaluation framework.} All results are based on the RIKER methodology. While RIKER's ground-truth-first approach avoids many pitfalls of traditional evaluation (see Section~\ref{sec:methodology}), our findings have not been independently replicated with alternative frameworks.

\textbf{English only.} All generated documents and questions are in English. Fabrication rates and context degradation patterns may differ for other languages, particularly for models with varying multilingual training data proportions.

\textbf{Open-weight models only.} Our evaluation covers 35 open-weight models served via vLLM. Proprietary API-based models (GPT-4, Claude, Gemini) were not tested and may exhibit different fabrication profiles.

\textbf{Document Q\&A focus.} Our evaluation specifically targets document-grounded question answering. Fabrication rates and temperature effects may differ for other LLM applications such as summarization, creative writing, or code generation.

\textbf{Temperature range.} We tested four temperature settings (0.0, 0.4, 0.7, 1.0). The gap between 0.0 and 0.4 is the largest in our range, and interesting dynamics may exist at intermediate values (e.g., 0.1, 0.2) that our coarser sampling does not capture.

\subsection{Future Work}\label{sec:future-work}

Several directions follow naturally from these findings.

\textbf{Finer temperature granularity.} The monotonic relationship between temperature and coherence loss, combined with the large 0.0-to-0.4 gap, suggests that finer sampling in the 0.0--0.4 range could identify optimal temperature sweet spots that balance accuracy, fabrication, and coherence.

\textbf{Longer context evaluation.} Our 200K evaluation already reveals dramatic degradation, but many models now advertise 1M+ token context windows. Extending RIKER evaluations to 400K and beyond would test whether the degradation trends we observe continue, plateau, or accelerate.

\textbf{Repetition penalty as an alternative to temperature.} If coherence loss at T=0.0 is driven by repetitive generation loops, repetition penalty parameters may provide an alternative mechanism to prevent these loops without the accuracy trade-offs of higher temperature. This deserves systematic evaluation.

\textbf{Fabrication-specific training interventions.} The strong model-family effect on fabrication rates (Section~\ref{sec:grounding-vs-fab}) suggests that targeted training interventions could substantially reduce fabrication. Investigating what training approaches produce the low fabrication rates seen in the GLM and MiniMax families could benefit the broader field.

\textbf{Multilingual evaluation.} Extending the RIKER methodology to non-English languages would test whether fabrication resistance transfers across languages or is language-specific.

\section{Conclusion}\label{sec:conclusion}

This paper set out to answer a deceptively simple question: \textit{how much do LLMs hallucinate when answering questions about documents they have been given?} Evaluating 35 open-weight models across three context lengths (32K, 128K, 200K), four temperatures, and three hardware platforms---consuming 172 billion tokens across more than 4,000 runs---we find that the answer is ``substantially, and unavoidably.'' Even under optimal conditions---best model, best temperature, temperature chosen specifically to minimize fabrication---the floor is non-zero and rises steeply with context length. At 32K, the best model (GLM 4.5) fabricates 1.19\% of answers, top-tier models fabricate 5--7\%, and the median model fabricates roughly 25\%. At 128K, the floor nearly triples to 3.19\% and only 5 of 26 tested models remain below 10\% fabrication. At 200K, no model stays below 10\%. In practice, fabrication rates will be higher still, since these numbers reflect each model's individually optimal temperature for minimizing fabrication---a setting practitioners are unlikely to know without dedicated benchmarking, and one that often trades off against overall accuracy.

Three findings carry direct implications for deployment. First, \textbf{model selection dominates all other factors}. Overall accuracy spans a 72-percentage-point range across models, and model family predicts fabrication resistance better than model size---GLM 4.5 fabricates at 1.19\% while models with far more parameters exceed 25\%. Second, \textbf{context length is the primary degradation driver}. Models that perform well at 32K can lose 10--30 percentage points of overall accuracy at 128K, with fabrication rates rising sharply. The context window a model advertises is not the context window it can use reliably. Third, \textbf{temperature requires nuance, not dogma}. While T=0.0 yields the best overall accuracy for roughly 60\% of model--context combinations, higher temperatures actually reduce fabrication rates for the majority of models, and T=0.0 dramatically increases the risk of coherence loss (infinite generation loops) at longer contexts---in extreme cases producing a 48$\times$ higher loop rate than T=1.0.

Beyond these practical findings, we identify a fundamental evaluation gap: \textbf{grounding capability and fabrication resistance are distinct, weakly correlated properties}. Models that excel at locating relevant information in documents may nonetheless fabricate answers at high rates. This decoupling means that retrieval-focused benchmarks---which dominate current long-context evaluation---are insufficient for assessing trustworthiness. Evaluation methodologies must test fabrication directly, using ground-truth-first approaches like RIKER that can distinguish genuine document-grounded answers from plausible but invented ones.

The scale of this study---35 models, 172 billion tokens, three hardware platforms---establishes baseline fabrication rates for the current generation of open-weight LLMs available during the experiment period (Christmas 2025). As models evolve, these baselines provide a quantitative reference point against which progress can be measured. The persistent gap between grounding capability and fabrication resistance suggests that reducing hallucination will require targeted training interventions, not simply greater scale.

\section*{Data Availability}
The experiment data, including all of the generated ground truth, document corpora, test sets, and the various model raw results, will be made available at \url{https://docs.kamiwaza.ai/research/datasets}.

\section*{Acknowledgments}

This research was made possible through the generous provision of GPU compute by  Signal65, who provided four servers with 8x AMD MI300X GPUs and two servers with 8xH200 GPUs for experimental evaluation. We thank Ryan Shrout, Brian Martin, Mitch Lewis, and Russ Fellows for their support. (\url{https://signal65.com/})

\section*{AI Usage Disclosure}

The researchers used the following generative AI services to assist with the manuscript:

\begin{itemize}
    \item Claude Code: Claude Opus 4.6
    \item Gemini 3 Pro Image (Nano Banana Pro)
\end{itemize}

    Most of the language in this paper was drafted by generative AI using RIKER project documents and code, plus certain reference material (especially previous work in PICARD and the original RIKER paper) and the raw and summarized RIKER experiment results in CSV form, and then heavily revised, edited and polished by human researchers. 100\% of this document was read and reviewed several times by human researchers.

In addition, all tables were created through generative AI directly using raw source data, and then reviewed by human researchers. Nano Banana Pro was used to generate the RIKER diagram by feeding it the final draft of this paper.

In all cases, final editorial control, technical validation, and intellectual responsibility rest solely with the human authors. The authors take full responsibility for the accuracy and integrity of all content in this manuscript.

\bibliography{references}

\end{document}